\documentclass[10pt,twocolumn,letterpaper]{article}

\usepackage[pagenumbers]{iccv} 
\usepackage{multirow}
\PassOptionsToPackage{table}{xcolor}
\usepackage{colortbl} 
\usepackage{graphicx}
\usepackage{caption}
\usepackage{amsfonts}
\usepackage{appendix}
\usepackage[accsupp]{axessibility}
\definecolor{iccvblue}{rgb}{0.21,0.49,0.74}
\definecolor{t2green}{rgb}{0.06,0.71,0.16}
\definecolor{t1blue}{rgb}{0.18,0.38,0.72}
\usepackage[pagebackref,breaklinks,colorlinks,allcolors=iccvblue]{hyperref}

\def\paperID{13474} 
\def\confName{ICCV}
\def\confYear{2025}

\title{Enhancing Mamba Decoder with Bidirectional Interaction in Multi-Task Dense Prediction}

\author{
    Mang Cao$^{1}$ \quad
    Sanping Zhou$^{1}$\thanks{Corresponding author.} \quad
    Yizhe Li$^{1}$ \quad
    Ye Deng$^{2}$ \quad
    Wenli Huang$^{3}$ \quad
    Le Wang$^{1}$
    \\[2mm] 
    $^{1}$National Key Laboratory of Human-Machine Hybrid Augmented Intelligence, \\
    National Engineering Research Center for Visual Information and Applications, \\
    Institute of Artificial Intelligence and Robotics, Xi’an Jiaotong University\\
    $^{2}$ School of Computing and Artificial Intelligence, Southwestern University of Finance and Economics \\
    $^{3}$	School of Electronic and Information Engineering, Ningbo University of Technology
    \\[2mm] 
    {\tt\small 3298592685@stu.xjtu.edu.cn, spzhou@xjtu.edu.cn}\\ 
}

\begin{document}
\maketitle
\begin{abstract}

Sufficient cross-task interaction is crucial for success in multi-task dense prediction. 
However, sufficient interaction often results in high computational complexity, forcing existing methods to face the trade-off between interaction completeness and computational efficiency.
To address this limitation, this work proposes a Bidirectional Interaction Mamba (BIM), which incorporates novel scanning mechanisms to adapt the Mamba modeling approach for multi-task dense prediction. 
On the one hand, we introduce a novel Bidirectional Interaction Scan (BI-Scan) mechanism, which constructs task-specific representations as bidirectional sequences during interaction. By integrating task-first and position-first scanning modes within a unified linear complexity architecture, BI-Scan efficiently preserves critical cross-task information.
On the other hand, we employ a Multi-Scale Scan~(MS-Scan) mechanism to achieve multi-granularity scene modeling. This design not only meets the diverse granularity requirements of various tasks but also enhances nuanced cross-task feature interactions. Extensive experiments on two challenging benchmarks, \emph{i.e.}, NYUD-V2 and PASCAL-Context, show the superiority of our BIM vs its state-of-the-art competitors. Codes are available online: \url{https://github.com/mmm-cc/BIM\_for\_MTL}.

\end{abstract}    
\section{Introduction}
\label{sec:intro}
Multi-task dense prediction is a critical visual task designed to simultaneously predict outputs for various pixel-level tasks, such as semantic segmentation, depth prediction, surface normal estimation, and saliency detection.
Existing approaches made mainly efforts~\cite{vandenhende2021multi,DBLP:journals/corr/abs-2009-09796} to refine task-general representations, obtained from the encoder, into task-specific representations. In these approaches, cross-task interaction has been proven to be significant in multi-task dense prediction~\cite{xu2018pad,zhang2023rethinking,ye2022inverted,sinodinos2024ema}, as it can enhance the robustness and effectiveness of task representations. 
\begin{figure}[t]
\begin{center}
   \includegraphics[width=\linewidth]{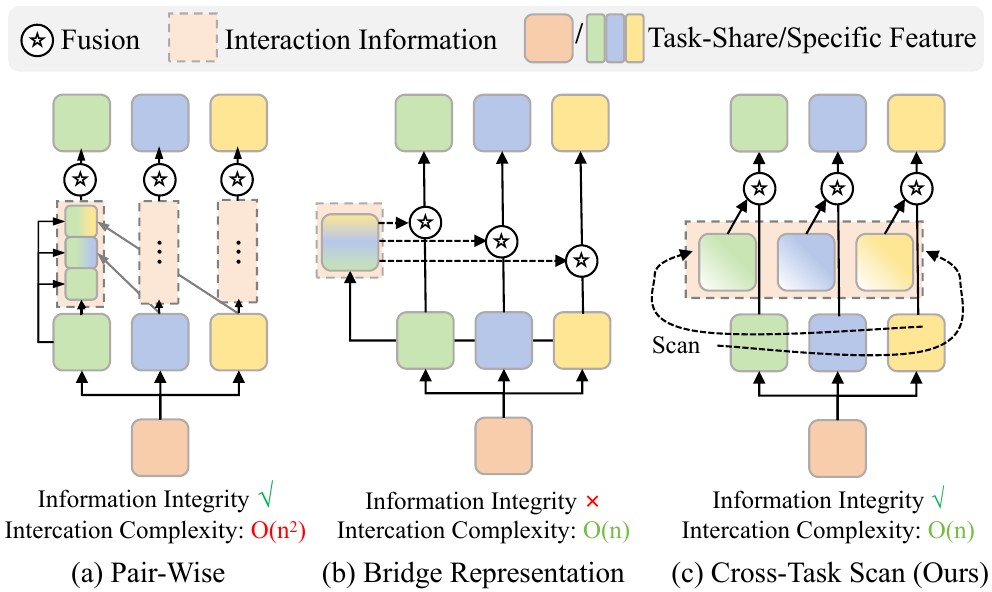}
\end{center}
   \caption{Illustration of different cross-task interaction strategies. (a) Pair-Wise~\cite{xu2018pad,gao2019nddr,sun2021task}: each task can access the complete information of other tasks, while the interaction complexity is $O(n^2)$. (b) Bridge Representation~\cite{shoouri2023efficient,bhattacharjee2022mult,zhang2023rethinking,lin2024mtmamba}: the interaction complexity is $O(n)$, yet, each task is limited to accessing only a partial subset of information from other tasks during interaction, which inevitably results in the loss of valuable details. (c) Ours Cross-Task Scan: It preserves the integrity of task information during interaction, with an interaction complexity of $O(n)$.}


\label{intro3}
\end{figure}

Initially, multi-task dense prediction methods primarily utilized Convolutional Neural Networks~(CNNs) with pair-wise interactions, as illustrated in \cref{intro3} (a). These networks~\cite{xu2018pad,gao2019nddr,sun2021task}
were carefully designed with specialized branches for each task, augmented by interaction modules that facilitated information exchanges between each pair of tasks.
These pair-wise interaction approaches maintain the integrity of task information during interaction, promoting fine-grained cross-task interaction.
While, the inherently limited receptive fields of CNN architectures often led to insufficient spatial interactions. In response to these challenges, 
transformer-style networks~\cite{ye2022inverted} demonstrate remarkable capability in modeling long-range dependencies. This capability substantially improves the interaction effectiveness of models in handling multi-task scenarios, leading to enhanced performance.
However, these sufficient cross-task interactions usually introduce high interaction complexity, which typically increases quadratically with the number of tasks. The high interaction complexity limits their applicability in certain powerful models, particularly transformers.



To enhance the efficiency of cross-task interactions, a crowd of methods~\cite{shoouri2023efficient,bhattacharjee2022mult,zhang2023rethinking,lin2024mtmamba} with linear interaction complexity has been proposed, as depicted in \cref{intro3} (b).
Typically, these methods strategically select a representative task~\cite{shoouri2023efficient,bhattacharjee2022mult} or extract representative information from all tasks~\cite{zhang2023rethinking,lin2024mtmamba} to construct a compact bridge representation that aims to preserve essential interaction information.
However, in these methods, the necessity of maintaining linear interaction complexity inherently limits the representational capacity of these bridge representations. 
This information compression inevitably leads to the degradation of task features, potentially discarding critical interaction details and ultimately resulting in suboptimal performance.
Therefore, how to enhance sufficient interaction capability while maintaining linear interaction complexity remains an unresolved issue.

In response to these challenges, based on Mamba architecture, we propose a simple yet effective Bidirectional Interaction Mamba (BIM) method that achieves comprehensive preservation of task information during interaction while maintaining linear interaction complexity.
Specifically, we introduce a Bidirectional Interaction Scan~(BI-Scan) mechanism to achieve sufficient cross-task interaction and a Multi-Scale Scan~(MS-Scan) mechanism to deliver a comprehensive visual representation. The BI-Scan mechanism constructs each task representation into sequences and deploys a bidirectional scanning strategy for modeling cross-task interactions, efficiently mitigating information loss while preserving linear interaction complexity. Furthermore, BI-Scan enables multi-level cross-task interactions by integrating task-first and position-first scanning modes, facilitating more nuanced modeling of task relationships.
In the MS-Scan, features are initially partitioned into multiple spaces, where the scene structure of images is modeled at various scales. Specific tasks subsequently integrate multi-scale scene structural information to enhance task-specific representations. During cross-task interaction, different task branches adaptively integrate interaction representations from multiple scales, addressing the varying granularity demands of each task for scene structural information.

The main contributions are summarized as follows:
\begin{itemize}
    \item We propose a simple yet effective Bidricational Interaction Mamba~(BIM) for multi-task dense prediction, enabling comprehensive cross-task interaction while preserving linear interaction complexity.
    \item We introduce a novel Bidirectional Interaction Scan~(BI-Scan) mechanism to maintain the integrity of task information during interactions. The BI-Scan incorporates two scanning modes: task-first mode and position-first mode, facilitating multi-level cross-task interactions.
    \item We design a Multi-Scale Scan (MS-Scan) mechanism to capture scene structural information at multiple scales, thereby addressing the diverse granularity requirements of various tasks and enhancing more detailed cross-task interactions in multi-task dense prediction.

\end{itemize}



\begin{figure*}[t]
  \centering
   \includegraphics[width=\linewidth]{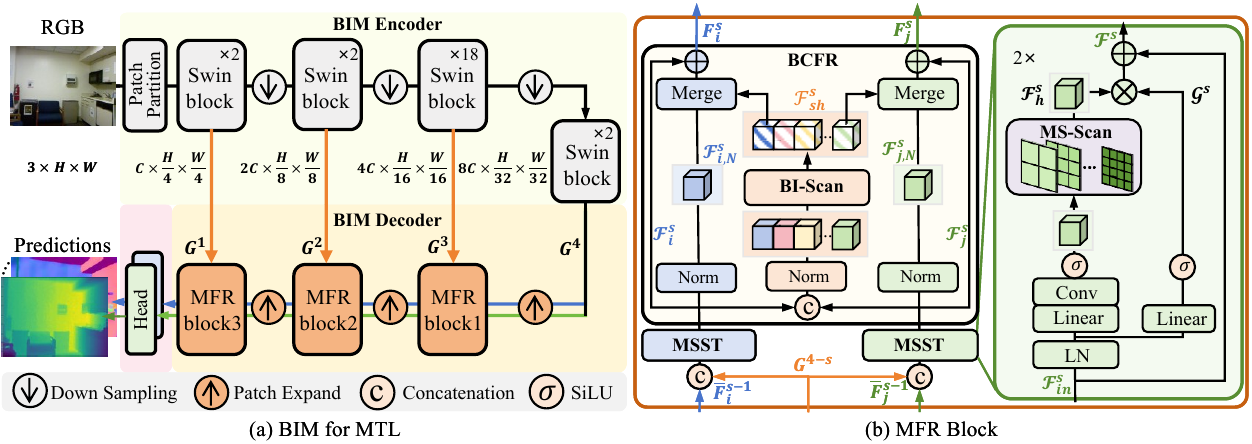}
   \caption{Framework of the proposed BIM for multi-task dense prediction. (a) Overview of BIM, illustrating with depth estimation and surface normal estimation tasks. (b) Details of the MFR block, including T task-specific MSST blocks and a task-shared BCFR block.}
\label{pip}
\end{figure*}


\section{Related Work}
\textbf{Multi-Task Learning. }Most existing multi-task learning works focused on training optimization and network design. Optimization approaches balance tasks via gradient manipulation~\citep{yu2020gradient,navon2022multi,jeong2024quantifying,ye2024adaptive} or loss-balancing~\citep{chen2018gradnorm,kendall2018multi} to coordinate resource allocation for various tasks during training. The approaches of network design 
seek to enhance task representation learning by developing diverse mechanisms.
Some CNN-based methods~\citep{xu2018pad,vandenhende2020mti} manually design interaction mechanisms to extract useful information across tasks.
With the advancement of Transformer, current methods have gained improved global task interaction capabilities to enhance task representation efficiency. They enable superior pairwise~\citep{bhattacharjee2022mult,shoouri2023efficient} or global~\citep{ye2022inverted,xu2023multi,ye2023taskprompter,li2024multitask} task interactions.
To mitigate computational complexity, MTMamba~\citep{lin2024mtmamba} introduce Mamba~\citep{gu2024mamba} to multi-task learning, which showcases effective long sequence modeling capabilities and achieving satisfactory performance. However, achieving both efficient and sufficient cross-task interaction remains a challenge.

\noindent \textbf{State Space Models.}
In efficient long-range dependency modeling methods, state space models~\citep{gu2021combining,DBLP:journals/corr/abs-2208-04933} has become a striking alternative to Transformers. \citep{DBLP:journals/corr/abs-2111-00396} proposed a Structured State Space Sequence~(S4) model based on a new parameterization, which alleviates the computational and memory efficiency issues faced by SSM.
Subsequently, numerous efforts~\citep{fu2023hungry,mehta2022long} are dedicated to bridging the performance disparity between SSMs and Transformers.
Mamba~\citep{gu2024mamba} introduced an input-based parameterization method and hardware-aware algorithm, achieving performance on par with Transformers in natural language processing. This success inspired various efforts~\citep{DBLP:conf/icml/ZhuL0W0W24} towards Mamba's adaptation for visual tasks.

The preservation of comprehensive image structural information poses a critical challenge in Mamba's sequential processing model, which has attracted considerable attention and effort~\citep{DBLP:conf/icml/ZhuL0W0W24,DBLP:journals/corr/abs-2401-10166,DBLP:journals/corr/abs-2403-17695,huang2024localmamba,zhao2024rs}. VMamba~\citep{DBLP:journals/corr/abs-2401-10166} proposes the 2D Selective Scan (SS2D), a four-way scanning mechanism tailored for spatial domain traversal, aimed at enhancing Mamba's image modeling capabilities. Subsequent research studies have explored various scan patterns and combinations tailored to different tasks or scenarios~\citep{DBLP:journals/corr/abs-2403-17695,huang2024localmamba,zhao2024rs}.
However, by utilizing fixed token sizes, these methods overlook the importance of hierarchical spatial structural information in visual tasks.
\section{Method}
\label{sec:method}

We first outline the overall architecture of BIM for multi-task dense prediction in Section 3.1, Subsequently, Section 3.2 focuses on the BIM Encoder, while Section 3.3 examines the BIM Decoder and its two scanning mechanisms: BI-Scan and MS-Scan. Finally, Section 3.4 discusses the optimization objectives.
\subsection{Framework Overview}
As shown~\cref{pip} (a), the proposed BIM framework consists of three main components: a task-shared encoder $\Phi$ for extracting task-generic representations, a Mamba decoder $\Theta$ for refining task features and T task-specific heads $\{H_t\}_{t=1}^T$ generating predictions $\{\hat{\textbf{Y}}_{t}\}_{t=1}^T$ for individual tasks. This can be formulated as:
\begin{equation}
\begin{aligned}
    \{F_1,F_2,\ldots,F_T\}=\Theta 
    (\Phi(\textnormal{I})),\\
    \hat{\textbf{Y}}_{t}=H_t(F_t)=\mathcal{P}_t ( \Psi(F_{t})),
\end{aligned}
\end{equation}
where $\textnormal{I}\in \mathbb{R}^{ 3 \times H \times W}$ denotes the RGB input image, $F_t$ denotes the features generated by the decoder, corresponding to the distinct task $t$. $\hat{\textbf{Y}}_{t}$ represents the prediction for task $t$, which has the same height $H$ and width $W$ as $\textnormal{I}$. $\Psi$ denotes a module designed to expand the task feature resolution to $H\times W$, which consists of a linear projection and a reshape operation. $\mathcal{P}_t$ represents a linear layer that projects the feature channels to a specific number required by the corresponding task.

\subsection{BIM Encoder}
The encoder shares similarities with other methods~\cite{lin2024mtmamba}. We utilized a pretrained Swin Transformer~\cite{liu2021swin} to extract task-generic features, which begins by dividing the input image $\textnormal{I}\in \mathbb{R}^{H \times W \times 3}$ into $\frac{H}{w} \times \frac{W}{w}$ tokens of dimension $C$ through patch partitioning and linear layers, where $w$ denotes the partition size.
These tokens are then processed through multiple stages involving alternating patch merging and Swin Transformer block processing, ultimately yielding hierarchical image representations:
\begin{equation}
    \textbf{G}=\{\textbf{G}^1, \textbf{G}^2, \textbf{G}^3,\textbf{G}^4\}=\Phi(\textnormal{I}),
\end{equation}
where $ \textbf{G}  $ represents the task-generic features extracted from the encoder $\Phi$. In our practical implementation, we utilize a partition size of 4, resulting in the following shapes for $\textbf{G}$: $C\times \frac{H}{4} \times \frac{W}{4}, 2C\times \frac{H}{8} \times \frac{W}{8}, 4C\times \frac{H}{16} \times \frac{W}{16}$ and $8C\times \frac{H}{32} \times \frac{W}{32}$, respectively.
\subsection{BIM Decoder}
The BIM Decoder consists of three Mamba Feature Refinement~(MFR) blocks, as depicted in \cref{pip} (a). These MFRs are designed to bridge the gap between task-generic and task-specific representations. They generate task representations $\{\mathbf{F}_1^{i},\mathbf{F}_2^{i},\ldots,\mathbf{F}_T^{i}\}_{i=1}^{3} $ corresponding to the first three encoder stages, with dimensions of $4C\times \frac{H}{16} \times \frac{W}{16}, 2C\times \frac{H}{8} \times \frac{W}{8}$, and $C\times \frac{H}{4} \times \frac{W}{4}$, respectively. Finally, the last refined features $\{\textbf{F}_{t}^{3} \}_{t=1}^{T}$ are input into the task heads as $\{F_t\}_{t=1}^T$ to generate the final predictions $\{\hat{\textbf{Y}}_{t}\}_{t=1}^{T}$.

We introduce two specialized Mamba blocks, Bidirectional Cross-Task Feature Refine~(BCFR) and Multi-Scale Single-task~(MSST) in each MFR block, as illustrated in \cref{pip} (b). 
The BCFR Block is designed to facilitate effective cross-task interaction while maintaining the integrity of task information. It begins by concatenating features $\{ \mathcal{F}^{s}_{t}\}_{t=1}^{T}$ from different task branches to construct task-shared features $\mathcal{F}_{sh}^{s}$ using the BI-Scan mechanism.
Subsequently, each task branch adaptively merges $\mathcal{F}_{sh}^{s}$ by a selection value $\mathcal{G}^{s}$ to obtain fine-tuned task representations $\textbf{F}^{s} = \{\textbf{F}^{s}_{t}\}_{t=1}^{T}$:
\begin{equation}
\textbf{F}_t^s=\mathcal{F}^{s}_{t}+\mathcal{G}^{s}_{t}\times\mathcal{F}^{s}_{sh}[t]+(1-\mathcal{G}^{s}_{t})\times \mathcal{F}^{s}_{t,N},
\end{equation}
where $\mathcal{F}^{s}_{t,N} $ is is derived from $\mathcal{F}^{s}_{t}$ through layer normalization, and $\mathcal{G}^{s}_{t}$ is obtained from $\mathcal{F}^{s}_{t,N}$ via a linear layer followed by a Sigmoid function. $\mathcal{F}^{s}_{sh}[t]$ is the $t$-th component of  $\mathcal{F}^{s}_{sh}$ corresponding to the $t$-th task.
The MSST primarily aims to construct comprehensive representations through task-internal interactions. Its architecture is illustrated in \cref{pip} (b) and comprises two main branches: the scan branch and the gating branch. In the scan branch, we integrate a hierarchical global scene structure representation $\mathcal{F}_{h}^{s}$ using a novel MS-Scan mechanism. Simultaneously, a gating signal $\mathcal{G}^{s}$ from the gate branch regulates the flow of information. Subsequently, we adjust the channel dimensions of the multi-scale scene representation by applying a linear projection $\mathcal{P}$ and establish a residual connection between $\mathcal{F}_{h}^{s}$ and the input $\mathcal{F}_{in}^{s}$. This process is repeated twice to produce the final output $\mathcal{F}^{s}$:
\begin{equation}
    \mathcal{F}^{s}= \mathcal{F}_{in}^{s} + \mathcal{P}(\mathcal{F}_{h}^{s} \times \mathcal{G}^{s}).
\end{equation}

\noindent \textbf{Bidirectional Interaction Scan.}
\label{sec_bct}
\begin{figure}[t]
  \centering
   \includegraphics[width=\linewidth]{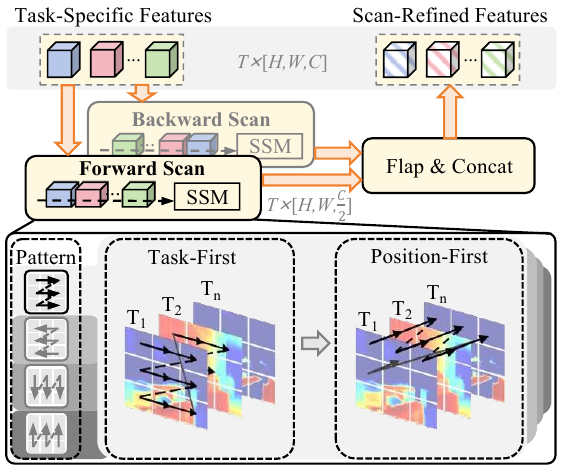}
   \caption{Framework of BI-Scan. It consists of two modules, Forward Scan and Backward Scan, each including Task-First and Position-First cross-task scanning modes.}
   \label{bct}
\end{figure}
To maintain task information integrity during interaction with linear complexity, we propose a novel BI-Scan that serves as a cornerstone of BIM, as shown in \cref{bct}. This module takes the representations of different tasks $\{\mathcal{F}_{t,N}^{s}\}_{t=1}^T$ as input, which is initially partitioned along the channel dimension into dual-branch representations, each with the dimensions $H\times W \times \frac{C}{2}$. The former undergoes processing through our Forward Scan module, while the latter is reversed along the task dimension before the Backward Scan processing. Specifically, the Forward Scan architecture implements hierarchical cross-task interaction through two scanning modes: Task-First mode and Position-First mode. 

\textit{Task-First Mode}: Following a similar configuration to SS2D in Vmamba~\cite{DBLP:journals/corr/abs-2401-10166}, we implement multiple distinct Scan Patterns for each scanning mode to facilitate enhanced task interaction. Without loss of generality, we use the first scan pattern as an example. 
As shown in \cref{bct}, in the task-first mode, the construction of global task information contains two processes: (\textbf{1}) Pattern-guided feature serialization ($\mathcal{U}_1$): each task feature is serialized according to the scan pattern, resulting in $T$ subsequences, each of length $H\times W$. (\textbf{2}) Task order-guided subsequence aggregation ($\mathcal{O}_1$): these subsequences are concatenated along the length dimension according to the specified task order, producing a global sequence $S$ of length $T\times (H \times W)$. These processes can be expressed as:
\begin{equation}
    S_1=\mathcal{O}_1(\mathcal{U}_1(\mathcal{F}_t^{s},\mathcal{F}_2^{s},\ldots,\mathcal{F}_T^{s})) \in \mathbb{R}^{(T\times (H \times W)) \times \frac{C}{2}}.
\end{equation}
This sequence undergoes global modeling through SSM~\cite{gu2024mamba}, yielding task-level interaction information.

\textit{Position-First Mode}: In the position-first mode, the output of Task-First mode is first restored from the sequence to the input task feature size, recorded as $\{\bar{F}_t^s\}_{t=1}^T \in \mathbb{R}^{H\times W\times \frac{C}{2}}$. Subsequently, the construction of interaction representations also involves two steps: (\textbf{1}) Task order-guided feature serialization ($\mathcal{O}_2$): Spatially aligned task representation tokens are constructed into $H\times W$subsequences, each of length T, according to the specified task order. (\textbf{2}) Pattern-guided subsequence aggregation ($\mathcal{U}_2$): These subsequences are concatenated along the length dimension according to the scan pattern, resulting in a sequence of length $ (H \times W)\times T$. The above processes can be expressed as:
\begin{equation}
S_2=\mathcal{U}_2(\mathcal{O}_2(\mathcal{\bar{F}}_1^{s},\mathcal{\bar{F}}_2^{s},\ldots,\mathcal{\bar{F}}_T^{s})) \in \mathbb{R}^{((H \times W)\times T) \times \frac{C}{2}}.
\end{equation}
This sequence is also processed through SSM, allowing for the derivation of task-level interaction insights.
 

Subsequently, the outputs corresponding to all patterns are fused via element-wise summation, yielding the output of Forward Scan, denoted as $\mathcal{F}_{for}^s$. Finally, the output of the Backward Scan, $\mathcal{F}_{back}^s$, is reversed along the task dimension and concatenated with $\mathcal{F}_{for}^s$ to yield the refined task representations $\mathcal{F}_{sh}^{s}$. In practical application, when employing BI-Scan independently, the selection of Scan Patterns aligns with those used in SS2D~\cite{DBLP:journals/corr/abs-2401-10166}, specifically, the four directions shown in \cref{bct}. When MS-Scan is introduced within the BI-Scan module, we incorporate Scan Patterns of varying scales without increasing computational complexity.

\noindent\textbf{Multi-Scale Scan.}
\label{sec_msscan}
\begin{figure}
  \centering
   \includegraphics[width=\linewidth]{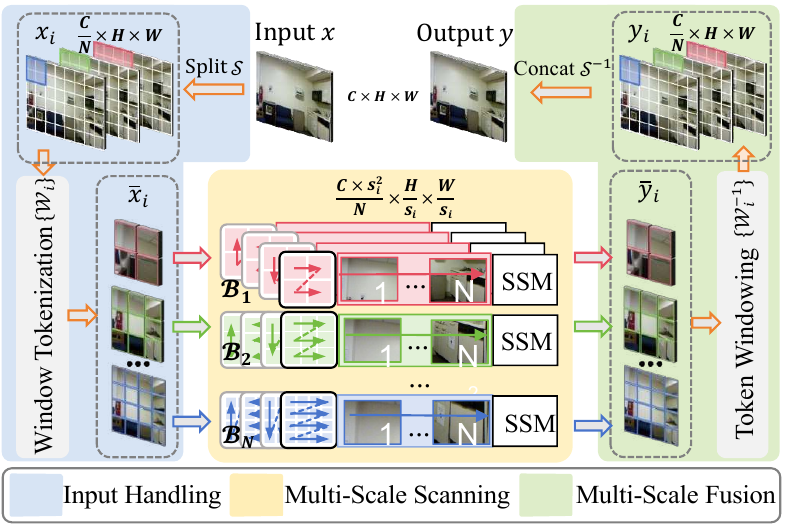}
   \caption{Instructions for MS-Scan. It consists of three distinct operations, Input Handling, Multi-Scale Scanning, and Multi-Scale Fusion. In Multi-Scale Scanning, each branch $\mathcal{B}_{i}$ has a distinct scanning size $s_i$.}
 \label{msscan}
\end{figure}
Mamba processes features by converting them into sequences of tokens. While a variety of tokenization strategies exist for text, image tokenization requires careful consideration of multiple approaches. Previous research has demonstrated that multi-scale processing is particularly effective for image data. To leverage this advantage, We propose the MS-Scan mechanism that requires minimal additional computational cost compared to single-scale scanning, as illustrated in \cref{msscan} (a). In this framework, we employ multiple scanning scales, denoted as $\{s_i\}_{i=1}^{N}$, in multiple branches $\{\mathcal{B}_{i}\}_{i=1}^{N}$, and transform the input image feature $x \in \mathbb{R}^{C\times H \times W}$ into token sequences of varying dimensions for Mamba modeling. For instance, the initial image feature can be tokenized into a sequence with a total length of $H \times W$, where each token has a dimension of $C$. When applying a scanning scale of $s_i = 2$, the image is divided into non-overlapping patches, resulting in a tokenized feature sequence with a dimensionality of $4C \times \left(\frac{H}{2} \times \frac{W}{2}\right)$. Specifically, MS-Scan comprises three key components: input handling, multi-scale scanning, and multi-scale fusion.

\textit{Input Handling}: To construct inputs for \(N\) different scanning branches $\mathcal{B}=\{\mathcal{B}_{i}\}_{i=1}^{N}$, we perform two main operations. \textbf{(1)} Channel Split ($\mathcal{S}$): We begin by splitting the input representation $x$ into $N$ sub-features \(\{x_{i}\}_{i=1}^{N} \in \mathbb{R}^{m \times H \times W}\) along the channel dimension, where $m = \frac{C}{N}$. 
\textbf{(2)} Window Tokenization ($\mathcal{W}_{i}$): 
For the $i$-th branch $\mathcal{B}_{i}$  with scanning scale $s_{i}$, we first divide $x_{i}$  into $\frac{H}{s_{i}} \times \frac{W}{s_{i}}$ non-overlapping patches, each of size $m \times s_{i} \times s_{i}$. Subsequently, we concatenate the pixel feature values in each patch along the channel dimension, resulting in the scan input $\bar{x}_{i}$ with a shape of $(m\times s_{i}\times s_{i}) \times \frac{H}{s_{i}} \times \frac{W}{s_{i}}$. Ultimately, we obtain the input for all branches:
\begin{equation}
    \{\bar{x}_{1}, \bar{x}_{2}, \ldots, \bar{x}_{N}\} = \{\mathcal{W}_{1}, \mathcal{W}_{2}, \ldots, \mathcal{W}_{N}\}( \mathcal{S}(x)).
\end{equation}

\textit{Multi-Scale Scanning}: Following the input handling process, we employ distinct scanning scales to construct a multi-scale scene representation in each branch. For all branches $\mathcal{B}$, we utilize the four-way scanning method~(SS2D) from VMamba~\citep{DBLP:journals/corr/abs-2401-10166} to generate scene features at the specified scale. This method creates four token sequences, each shaped as $C_{i}\times(H_i \times W_{i})$, by scanning the input features $\bar{x}_{i}\in \mathbb{R}^{C_i\times H_i\times W_i}$ in four directions. The resulting sequences are then processed by SSM~\citep{gu2024mamba} and combined to produce the output feature $\bar{y}=\{\bar{y}_{i}\}_{i=1}^{N}$:
\begin{equation}
    \bar{y}_{i} = \textnormal{SS2D}(\bar{x}_{i}) \in \mathbb{R}^{C_{i} \times H_{i} \times W_{i}}, i\in\{1,2,\ldots,N\}.
\end{equation}

\textit{Multi-Scale Fusion}: In our approach to multi-scale feature fusion, we adopt a methodology that reverses the input handling process, consisting of two key steps. \textbf{(1)} Token Windowing ($\mathcal{W}_{i}^{-1}$): For each branch \(\mathcal{B}_{i}\), we split each pixel feature into $s_{i}\times s_{i}$  segments along the channel dimension, resulting in a set of feature segments
$\{\bar{y}_{i,j}\}_{j=1}^{s_{i}\times s_{i}}\in \mathbb{R}^{m \times 1 \times 1}$.
These segments are then concatenated along the spatial dimensions (height and width) to form patches, which are subsequently combined to produce the output for \(\mathcal{B}_{i}\).
\textbf{(2)} Channel Concatenation ($\mathcal{S}^{-1}$): We concatenate the features from all branches along the channel dimension, yielding the final output feature $y \in \mathbb{R}^{C \times H \times W}$:


\begin{equation}
    y=  \mathcal{S}^{-1} (\mathcal{W}^{-1}_{1}(\bar{y}_{1}),\mathcal{W}_{2}^{-1}(\bar{y}_{2}),\ldots,\mathcal{W}^{-1}_{N}(\bar{y}_{N})),
\end{equation}
where $\mathcal{S}^{-1}$ and $\mathcal{W}_{i}^{-1}$ refer to the inverse operation of $\mathcal{S}$ and $\mathcal{W}_{i}$ respectively.

\subsection{Optimization Objective}
We jointly train all tasks to optimize BIM encoder $\Phi$, BIM decoder $\Theta$, and task heads $\{H\}_{i=1}^{T}$ simultaneously. To maintain consistency
with previous approaches, we use $L1$ loss for depth estimation and surface normal
estimation tasks and the cross-entropy loss for other tasks, therefore, the optimization objective can be expressed as follows:
\begin{equation}
    L = \sum_{t\in \mathcal{T}}\lambda_{t} \mathcal{L}_{t}(\hat{\textbf{Y}}_t ,\textbf{Y}_{t}),
\end{equation}
where $\mathcal{T}$ is the set of all tasks, $\lambda_{t}$, $\mathcal{L}_{t}$ and $\textbf{Y}_{t}$ are the loss weight, loss function, and task label for image $\textnormal{I}$ in task $t$, respectively.



\section{Experiment}
\label{sec:exp}
\subsection{Experimental Setup}
\textbf{Datasets.} We performed experiments using the benchmark datasets NYUD-v2~\cite{silberman2012indoor} and PASCAL-Context~\cite{chen2014detect}.
NYUD-v2 primarily focuses on indoor scenes, with 795 and 654 RGB images for training and testing purposes. Tasks in NYUD-v2 include semantic segmentation, monocular depth estimation, surface normal estimation, and object boundary detection.
PASCAL-Context encompasses indoor and outdoor scenes, offering pixel-level labels for tasks like semantic segmentation, human parsing, object boundary detection, surface normal estimation, and saliency detection tasks. The PASCAL-Context dataset contains 4,998 training images and 5,105 test images.

\noindent\textbf{Implementation Details.}
We employ a pretrained Swin-Large Transformer~\cite{liu2021swin} on ImageNet-22K~\cite{deng2009imagenet} as our encoder. Our models are trained on the NYUD-v2 dataset for 50,000 iterations with a batch size of 4, and on the Pascal Context dataset for 75,000 iterations with a batch size of 6. Across all datasets, we use the Adam optimizer with a learning rate of $5\times10^{-5}$ and a weight decay rate of $1\times10^{-5}$, alongside a polynomial learning rate scheduler. The preliminary decoder has an output channel number of 768. 
We follow common practice~\cite{ye2022inverted,lin2024mtmamba} in resizing the input images and applying data augmentation techniques. Specifically, we resize the input images of NYUD-v2 and PASCAL-Context to $448 \times 576$ and $512 \times 512$, respectively, and apply random color jittering, random cropping, random scaling, and random horizontal flipping.


\subsection{Comparisons with State-of-the-art Methods}

\noindent\textbf{Main Results.} \cref{tabnyud} and \cref{tabpscal} report a comparison of the proposed BIM against previous state-of-the-art methods, including MTmamba~\cite{lin2024mtmamba}, MQTransformer~\cite{xu2023multi}, TaskPrompter~\cite{ye2023taskprompter},InvPT++~\cite{ye2024invpt++},InvPT~\cite{ye2022inverted}, ATRC~\cite{bruggemann2021exploring}, MTI-Net~\cite{vandenhende2020mti}, PAD-Net~\cite{xu2018pad}, PSD~\cite{zhang2019pattern}, PAP~\cite{zhou2020pattern}, Cross-Stitch~\cite{misra2016cross} and ASTMT~\cite{maninis2019attentive} on NYUD-V2 and PASCAL-Context dataset respectively.
Notably, the previous best
method, \emph{i.e.}, MTmamba, and our BIM are built upon the Transformer-encoder Mamba-decoder architecture with the same backbone. On NYUD-v2, the performance of Semseg is clearly boosted from the previous best,
\emph{i.e.}, 55.82 to 57.40 (\textbf{+1.52}). On Pascal-Context, we achieved superior performance on all tasks compared to MTMamba.

\begin{table}[ht]
    \footnotesize
    \caption{Quantitative comparison with other state-of-the-art methods on NYUD-v2 dataset.}
     \label{tabnyud}
        \centering
        \setlength{\tabcolsep}{4pt}
     \resizebox{0.45\textwidth}{!}
     {
        \begin{tabular}{c c c c c}
                \toprule
                \multirow{2}{*}{\textbf{Model}}& \textbf{Semseg}& \textbf{Depth}& \textbf{Normal}& \textbf{Boundary}\\
                & mIoU $\uparrow$ & RMSE$\downarrow$ & mErr $\downarrow$ & odsF $\uparrow$ \\
                \midrule
                 \textbf{CNN based}&&&&\\
                Cross-Stitch\cite{misra2016cross}&36.34 &0.6290 & 20.88& 76.38\\
                PAP\cite{zhou2020pattern}&36.72 &0.6178 & 20.82& 76.42\\
                PSD\cite{zhang2019pattern}&36.69&0.6246 & 20.87& 76.42\\
                PAD-Net\cite{xu2018pad}&36.61&0.6270& 20.85& 76.38\\
                MTI-Net\cite{vandenhende2020mti}&45.94 &0.5365& 20.27& 77.86\\
                ATRC\cite{bruggemann2021exploring}&46.33 &0.5363& 20.18& 77.94\\
                \midrule
                \textbf{Transformer based}&&&&\\
                InvPT\cite{ye2022inverted}&53.66 &0.5183 &19.04& 78.10\\
                TaskPrompter\cite{ye2023taskprompter}& 55.30 &0.5152& \textbf{18.47}& 78.20\\
                InvPT++\cite{ye2024invpt++}&53.85& 0.5096 &18.67& 78.10\\
                MQTransformer\cite{xu2023multi}&54.84&0.5325 & 19.67& 78.20\\
                \midrule
                \textbf{Mamba based}&&&&\\
                MTMamba\cite{lin2024mtmamba}&\underline{55.82} &\underline{0.5066} & 18.63&\underline{78.70}\\
                 \rowcolor{gray!20}
                 BIM(Ours) & \textbf{57.40}& \textbf{0.4733}& \underline{18.55}&\textbf{78.72} \\
                \bottomrule
        \end{tabular}
        }
\end{table}

\begin{table}[ht]
        
        \footnotesize
        \caption{Quantitative comparison with other state-of-the-art methods on Pascal-Context dataset.}
        \centering
        \setlength{\tabcolsep}{2pt}
        \resizebox{0.475\textwidth}{!}{
            \begin{tabular}{c c c c c c}
                \toprule
                \multirow{2}{*}{\textbf{Model}}& \textbf{Semseg}& \textbf{Parsing}& \textbf{Saliency}& \textbf{Normal}& \textbf{Boundary}\\
                 & mIoU $\uparrow$ & mIoU $\uparrow$ & maxF$\uparrow$ & mErr $\downarrow$ & odsF $\uparrow$ \\
                \midrule
                \textbf{CNN based}&&&&&\\
                PAD-Net\cite{xu2018pad}&  53.60& 59.60&  65.80&15.30& 72.50 \\
                ASTMT\cite{maninis2019attentive}& 68.00& 61.10&  65.70& 14.70&72.40 \\
                MTI-Net\cite{vandenhende2020mti}&  61.70&  60.18&  84.78&14.23&70.80 \\
                ATRC\cite{bruggemann2021exploring}& 62.69&  59.42&  84.70& 14.20& 70.96 \\
                ATRC-ASPP\cite{bruggemann2021exploring}&  63.60&  60.23&  83.91& 14.30& 70.86 \\
                ATRC-BMTAS\cite{bruggemann2021exploring}&  67.67& 62.93&  82.29& 14.24& 72.42 \\
                \midrule
                \textbf{Transformer based}&&&&&\\
                InvPT\cite{ye2022inverted}& 79.03& 67.61& \underline{84.81}&14.15&73.00 \\
               
                TaskPrompter\cite{ye2023taskprompter}& 80.89& 68.89& \textbf{84.83}& \textbf{13.72}& 73.50\\
                 InvPT++\cite{ye2024invpt++}&80.22 &69.12& 84.74 &\underline{13.73}& 74.20\\
                MQTransformer\cite{xu2023multi}&  78.93& 67.41&83.58&14.21& 73.90 \\
                \midrule
                \textbf{Mamba based}&&&&&\\
                MTmamba\cite{lin2024mtmamba}& \underline{81.11}&\underline{72.62}& 84.14&14.14&\underline{78.80} \\
                \rowcolor{gray!20}
                BIM(Ours)&
                \textbf{81.25}& \textbf{73.15}& 84.14& 14.13&\textbf{79.05} \\
                \bottomrule
            \end{tabular}
        }
        \label{tabpscal}
    
\end{table}

\noindent\textbf{Comparisons on Model Size and Computational Cost.} In \cref{complexity}, we compare the FLOPs and parameter counts with other State-of-the-art methods. With the comparable computational cost, our BIM shows significant outperforms. our light-vision BIM~(DBIM), which implements dilated sampling in MS-Scan, shows outperformance with reduced computational cost and parameter numbers, showcasing the advantage of these designs. 
Specifically, in MS-Scan, we use dilated sampling in each scan branch $\mathcal{B}$ to lighten the operation of MS-Scan. Specifically: (1) We only sample one token in a window when generating multi-scale sequences from image features, instead of all tokens. (2) When restoring the sequence to image features, we perform linear interpolation between these selected tokens. These operations do not introduce any parameters, and the computational burden is reduced because only a subset of tokens are used for modeling.

\begin{table}[t]
\setlength{\tabcolsep}{2pt}
\footnotesize
\centering
\caption{Complexity comparisons on NYUD-v2. \dag denotes the result is from~\cite{lu2024swiss}.}
\label{complexity}
 \resizebox{0.475\textwidth}{!}{
\begin{tabular}{c| c c c c| c c}
\toprule
\multirow{2}{*}{\textbf{Model}}& \textbf{Semseg}& \textbf{Depth}& \textbf{Normal}& \textbf{Boundary} &\textbf{FLOPs}& \textbf{\# Params}\\ 

& mIoU $\uparrow$ & RMSE$\downarrow$ &mErr $\downarrow$ &odsF $\uparrow$ &(G)$\downarrow$ &  (M)$\downarrow$ \\
\midrule
InvPT\cite{ye2022inverted}&53.66 &0.5183 &19.04& 78.10&598&402\\

TaskPrompter\cite{ye2023taskprompter}& 55.30 &0.5152& \textbf{18.47}& 78.20& 470&392\\
InvPT++$^{\dag}$\cite{ye2024invpt++}&53.85& 0.5096 &18.67& 78.10&$-$&$\sim$402\\
MTMamba\cite{lin2024mtmamba}& 55.82 & 0.5066& 18.63& 78.70 &541&308\\

\rowcolor{gray!20}
DBIM (Ours)&  \underline{56.41}  &  \underline{0.4789} & 18.66 & \textbf{78.80} & 495& 290\\
\rowcolor{gray!20}
BIM (Ours) & \textbf{57.40}& \textbf{0.4733}& \underline{18.55}&\underline{78.72} & 547&388\\
\bottomrule
\end{tabular}
}
\end{table}

\begin{figure*}[ht]
\centering
\includegraphics[width=\textwidth]{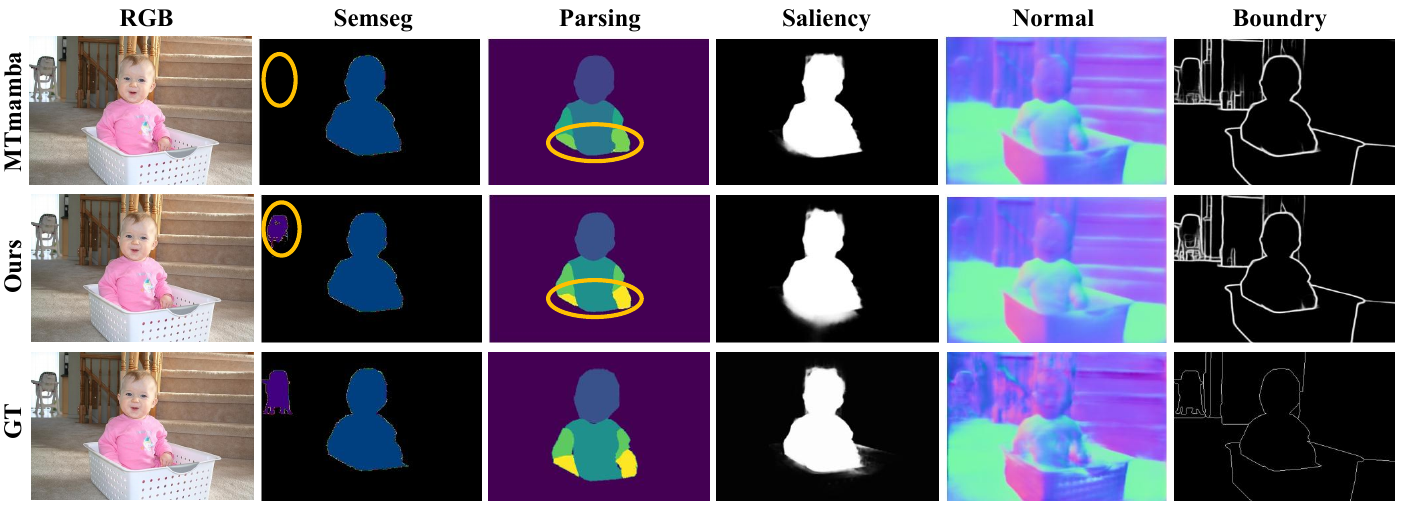}
\caption{Qualitative comparison with the best-performing method on Pascal-Context. Our method generates better details.} 
\label{vis1ny}
\end{figure*}

\noindent\textbf{Qualitative Comparisons.} We qualitatively compared our proposed BIM with the previous best-performing method, as shown in \cref{vis1ny}. Our method shows clear improvements in detail, as highlighted in the circled regions.

\subsection{Ablation Study}
\begin{figure}[t]
    \centering
        \includegraphics[width=0.485\textwidth]{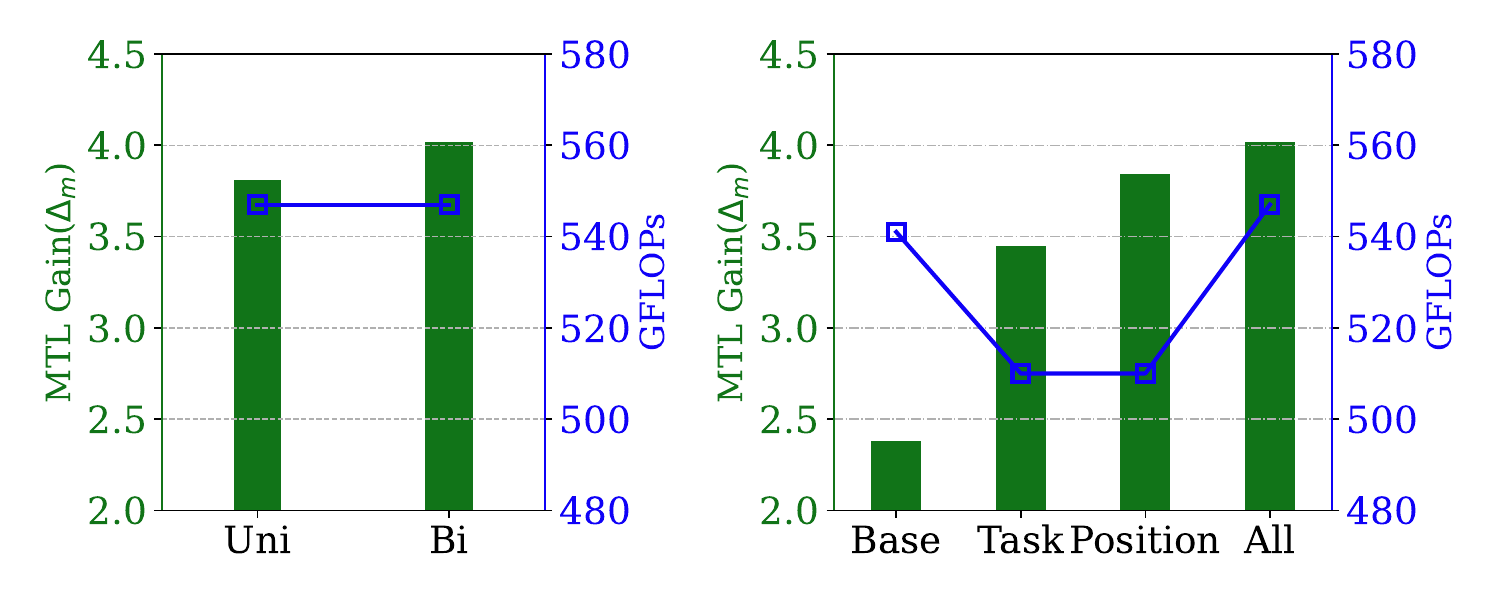} 
        \captionof{figure}{Ablation study on Bidirectional Scan (left) and two scanning modes (right) in BI-Scan.} 
        \label{ablationbct}
\end{figure}
\noindent\textbf{Effectiveness of Different Components.} We performed ablation experiments on BI-Scan and MS-Scan using the NYUD-V2 dataset. These experiments all used Swin-Large Transformer as encoder. We use the same STL and MTL base model as in MTMamba. (1) The term ``STL" indicates that each task is equipped with a separate network, utilizing two Swin Transformer blocks in each decoder stage, and (2) the ``MTL" indicates that all tasks share a common encoder. (3) ``Baseline" refers to MTMamba. (4) ``BI-Scan" indicates that the MS-Scan is replaced with SS2D throughout the entire model. (5) ``MS-Scan" refers to the implementation of the same cross-task interaction strategy as that used in the baseline. Lastly, (6) ``BIM" represents the default method.
As depicted in \cref{tabcomponents}, The BI-Scan improved model performance with fewer parameters, emphasizing the necessity of retaining task-specific details in cross-task interactions. Meanwhile, the MS-Scan enabled multi-granularity scene modeling, further enhancing performance.


\begin{table}[t]
    \centering
    \footnotesize
    \setlength{\tabcolsep}{1pt}
    \caption{Effectiveness of different components on NYUD-v2.}
        \label{tabcomponents}
        \resizebox{\linewidth}{!}{
            \begin{tabular}{c|c c c c c|c c}
                \toprule
                \multirow{2}{*}{\textbf{Model}} & \textbf{Semseg} & \textbf{Depth} & \textbf{Normal} & \textbf{Boundary} & \textbf{MTL Gain} & \textbf{FLOPs} & \textbf{\# Params} \\
                & mIoU $\uparrow$ & RMSE $\downarrow$ & mErr $\downarrow$ & odsF $\uparrow$ & $\Delta_{m} \uparrow$ & (G) $\downarrow$ & (M) $\downarrow$ \\
                \midrule
                STL & 54.32 & 0.5166 & 19.21 & 77.30 & +0.00 & 1075 & 889 \\
                MTL & 53.72 & 0.5239 & 19.97 & 76.50 & -1.87 & 466 & 303 \\
                \midrule
                Baseline & 55.82 & 0.5066 & 18.63 & 78.70 & +2.38 & 541 & 308 \\
                BI-Scan & 57.11 & 0.4856 & 18.66 & \textbf{78.90} & +4.02 & 547 & 290 \\
                MS-Scan & 57.40 & 0.4829 & 18.79 & 78.60 & +4.00 & 541 & 397 \\
                BIM & \textbf{57.40} & \textbf{0.4733} & \textbf{18.55} & 78.72 & \textbf{+4.83} & 547 & 388 \\
                \bottomrule
            \end{tabular}
            }
\end{table}

\noindent\textbf{Effectiveness of Bidirectional Scan in BI-Scan.} 
As shown in \cref{ablationbct} (left), we compared the performance of unidirectional (Uni) and bidirectional (Bi) cross-task scanning. The results show that the introduction of bidirectional scanning significantly enhances the cross-task modeling capability, thereby improving the overall performance of the model with only a negligible increase in computational cost.




\noindent\textbf{Effectiveness of Two Scan Modes in BI-Scan.} As shown in \cref{ablationbct} (right), we conducted ablation experiments on the Task-First Scan mode and Position-First Scan mode within BI-Scan. The results indicate that both Task-First~(Task) and Position-First~(Position) interaction modes achieve superior performance compared to the baseline~(Base) at a lower computational cost. Furthermore, employing both modes together~(All) results in further performance improvements. Ultimately, we incorporate both Task-First mode and Position-First mode as the final configuration of BI-Scan for BIM.


\noindent\textbf{Ablation on Task Order in BI-Scan.} We experiment on NYUD-v2 to study the effect of the task order during scan interaction. In addition to fixed task orders, we also implemented a random task order modeling approach~(Random), in which the order of tasks in BI-Scan is randomly shuffled during both training and testing. For clarity, we refer to the tasks Semseg, Depth, Normal, and Boundary as S, D, N, and B, respectively. As shown in \cref{tabtaskorder}, the overall performance of various fixed order models is relatively similar, while the Random setting exhibits a slight decrease in overall performance, which remains within an acceptable range. This indicates that our proposed BI-Scan demonstrates robustness to the task order. Ultimately, we adopt the first task order setting as the final configuration of BI-Scan for BIM.

\begin{table}[ht]
\caption{Ablation on task order on NYUD-v2.}
    \footnotesize  
    \centering
    \setlength{\tabcolsep}{5pt}
     \resizebox{0.48\textwidth}{!}{
        \begin{tabular}{c| c c c c c}
                \toprule
                
              \multirow{2}{*}{\textbf{Order}}& \textbf{Semseg}& \textbf{Depth}& \textbf{Normal}& \textbf{Boundary}&\textbf{MTL Gain}\\
                & mIoU $\uparrow$ & RMSE$\downarrow$ & mErr $\downarrow$ & odsF $\uparrow$ &$\Delta_{m} \uparrow$\\
                \midrule
            
               S-D-N-B& 57.11&0.4856&\textbf{18.66}&78.90& +4.02\\
               N-S-D-B& 56.15&\textbf{0.4832}&18.66	&\textbf{79.00}& +3.72\\
               B-S-D-N&56.97& 0.4836& 18.73&78.90 &+3.96\\
                S-N-D-B& \textbf{57.52}&0.4843&18.71&78.90& \textbf{+4.20}\\
                \midrule
               Random& 55.41&0.4861&18.69&78.90& +3.17\\
                \bottomrule
        \end{tabular}
        }
 \label{tabtaskorder}
\end{table}

\noindent\textbf{Ablation on Scan Scale and Number of MS-Scan.} We experimented with the impact of varying scanning scales and numbers on model performance, as shown in \cref{tabscanscale} and \cref{tabscannumber}. Experiments were conducted on the NYUD-v2 dataset with Swin-Large Transformer as encoder. We experimented with three different settings for scale ablation: all MFRs use \{1,2\}, \{1,3\}, and \{1,4\} two scanning scales, respectively. For number ablation, We compared three different settings, \textbf{Type 1}: all three MFRs use \{1\} scale; \textbf{Type 2}: all three MFRs use \{1,4\} two scales; \textbf{Type 3}: three MFRs use \{1,2\}, \{1,2,4\} and \{1,2,4,6\}, respectively.
The results showed that in multi-scale scanning, employing appropriate scanning scale partitions can effectively enhance overall performance. Significantly, all settings yield notable performance enhancements. Ultimately, we adopt $\{1,4\}$ scale setting for all MFRs as the final configuration.
\begin{table}[ht]
\caption{Different scanning scales of MS-Scan on NYUD-v2.}
    \footnotesize
    \setlength{\tabcolsep}{4pt}
    \centering
     \resizebox{0.475\textwidth}{!}{
        \begin{tabular}{c| c c c c c}
                \toprule
              \multirow{2}{*}{\textbf{Scan Scale}}& \textbf{Semseg}& \textbf{Depth}& \textbf{Normal}& \textbf{Boundary}&\textbf{MTL Gain}\\
                & mIoU $\uparrow$ & RMSE$\downarrow$ & mErr $\downarrow$ & odsF $\uparrow$ &$\Delta_{m} \uparrow$\\
                \midrule
               \{1,2\}& 57.39&0.4756&18.60&\textbf{78.80}& +4.68\\
               \{1,3\}& \textbf{57.53}&0.4739&18.68&78.50& +4.62\\
                \{1,4\}&57.40& \textbf{0.4733}& \textbf{18.55}&78.72 &\textbf{+4.83}\\
                \bottomrule
        \end{tabular}
        }
 \label{tabscanscale}
\end{table}

\begin{table}[ht]
\caption{Different scanning number of MS-Scan on NYUD-v2.}
    \footnotesize
    \setlength{\tabcolsep}{4pt}
    \centering
    \resizebox{0.475\textwidth}{!}{
         \begin{tabular}{c|c c c c  c}
            \toprule
            \multirow{2}{*}{\textbf{Scan Number}} & \textbf{Semseg}& \textbf{Depth}& \textbf{Normal} & \textbf{Boundary} & \textbf{MTL Gain} \\
            & mIoU $\uparrow$ & RMSE $\downarrow$ & mErr $\downarrow$ & odsF $\uparrow$ &
            $\Delta_{m} \uparrow$ \\
            \midrule
            Type 1 & 57.11 & 0.4856 & 18.66 & \textbf{78.90} & +4.02 \\
            Type 2 & 57.40 & \textbf{0.4733} & \textbf{18.55} & 78.72 &\textbf{+4.83}\\
            Type 3 & 56.76 & 0.4756 & 18.59 & 78.70 & +4.35\\
            \bottomrule
        \end{tabular}
    }
    \label{tabscannumber}
\end{table}

\noindent\textbf{Performance on Different Encoders.} We evaluate the effect of model size on experimental performance, as presented in \cref{tabencoder}. All experiments were conducted on the NYUD-v2 dataset.
We compare our method with the previous best-performing model, MTMamba, using two different encoders: Swin-Base Transformer (denoted as `-Base') and Swin-Large Transformer (denoted as `-Large').
The results suggest that models with greater capacities typically exhibit superior performance. Furthermore, our approach has demonstrated superior performance across all encoder.

\begin{table}[ht]
\caption{Performance on different encoders.}
    \footnotesize
    \centering
     \resizebox{0.465\textwidth}{!}{
         \begin{tabular}{c|c c c c}
            \toprule
            \multirow{2}{*}{\textbf{Model}} & \textbf{Semseg}& \textbf{Depth}& \textbf{Normal}& \textbf{Boundary} \\
            & mIoU $\uparrow$ & RMSE$\downarrow$ &mErr $\downarrow$ &odsF $\uparrow$ \\
            \midrule
            MTMamba-Base&  53.62 &0.5126& 19.28 & 77.70 \\
             \rowcolor{gray!20}
            BIM-Base& 54.32 &0.4915& 18.89 & 77.74 \\
            MTMamba-Large&  55.82 & 0.5066 &18.63 & 78.70\\
            \rowcolor{gray!20}
            BIM-Large&\textbf{57.40}& \textbf{0.4733}& \textbf{18.55}&\textbf{78.72} \\
        \bottomrule
        \end{tabular}
        }
 \label{tabencoder}
\end{table}


\section{Conclusion}
\label{sec:con}

We introduce the Bidirectional Interaction Mamba (BIM) framework, a simple yet effective solution designed to address the inherent conflict between adequacy and complexity in cross-task interactions for multi-task dense prediction. We first propose a Bidirectional Interaction Scan~(BI-Scan) mechanism, which constructs task information as bidirectional sequences to model cross-task interactions efficiently. By integrating task-first and position-first modes within linear complexity framework, BI-Scan preserves the integrity of task-specific information during interaction while maintaining computational efficiency. We further develop a Multi-Scale Scan (MS-Scan) mechanism, which captures scene structure information at multiple scales, alleviating the difficulty of feature learning and enhancing nuanced cross-task feature interactions. 
Extensive experiments show our method significantly enhances model performance.
\section{Acknowledgement}
This work was supported in part by National Natural Science Foundation of China under Grants 62088102, U24A20325 and 12326608, Key Research and Development Plan of Shaanxi Province under Grant 2024PT-ZCK-80, and the Public Welfare Research Program of Ningbo City under Grant 2024S063.
{
    \small
    \bibliographystyle{ieeenat_fullname}
    \bibliography{main}

\begin{thebibliography}{43}
\providecommand{\natexlab}[1]{#1}
\providecommand{\url}[1]{\texttt{#1}}
\expandafter\ifx\csname urlstyle\endcsname\relax
  \providecommand{\doi}[1]{doi: #1}\else
  \providecommand{\doi}{doi: \begingroup \urlstyle{rm}\Url}\fi

\bibitem[Bhattacharjee et~al.(2022)Bhattacharjee, Zhang, S{\"u}sstrunk, and Salzmann]{bhattacharjee2022mult}
Deblina Bhattacharjee, Tong Zhang, Sabine S{\"u}sstrunk, and Mathieu Salzmann.
\newblock Mult: An end-to-end multitask learning transformer.
\newblock In \emph{Proceedings of the IEEE/CVF Conference on Computer Vision and Pattern Recognition}, pages 12031--12041, 2022.

\bibitem[Br{\"u}ggemann et~al.(2021)Br{\"u}ggemann, Kanakis, Obukhov, Georgoulis, and Van~Gool]{bruggemann2021exploring}
David Br{\"u}ggemann, Menelaos Kanakis, Anton Obukhov, Stamatios Georgoulis, and Luc Van~Gool.
\newblock Exploring relational context for multi-task dense prediction.
\newblock In \emph{Proceedings of the IEEE/CVF international conference on computer vision}, pages 15869--15878, 2021.

\bibitem[Chen et~al.(2014)Chen, Mottaghi, Liu, Fidler, Urtasun, and Yuille]{chen2014detect}
Xianjie Chen, Roozbeh Mottaghi, Xiaobai Liu, Sanja Fidler, Raquel Urtasun, and Alan Yuille.
\newblock Detect what you can: Detecting and representing objects using holistic models and body parts.
\newblock In \emph{Proceedings of the IEEE conference on computer vision and pattern recognition}, pages 1971--1978, 2014.

\bibitem[Chen et~al.(2018)Chen, Badrinarayanan, Lee, and Rabinovich]{chen2018gradnorm}
Zhao Chen, Vijay Badrinarayanan, Chen-Yu Lee, and Andrew Rabinovich.
\newblock Gradnorm: Gradient normalization for adaptive loss balancing in deep multitask networks.
\newblock In \emph{International conference on machine learning}, pages 794--803. PMLR, 2018.

\bibitem[Crawshaw(2020)]{DBLP:journals/corr/abs-2009-09796}
Michael Crawshaw.
\newblock Multi-task learning with deep neural networks: A survey.
\newblock \emph{CoRR}, abs/2009.09796, 2020.

\bibitem[Deng et~al.(2009)Deng, Dong, Socher, Li, Li, and Fei-Fei]{deng2009imagenet}
Jia Deng, Wei Dong, Richard Socher, Li-Jia Li, Kai Li, and Li Fei-Fei.
\newblock Imagenet: A large-scale hierarchical image database.
\newblock In \emph{2009 IEEE conference on computer vision and pattern recognition}, pages 248--255. Ieee, 2009.

\bibitem[Fu et~al.(2023)Fu, Dao, Saab, Thomas, Rudra, and Re]{fu2023hungry}
Daniel~Y Fu, Tri Dao, Khaled~Kamal Saab, Armin~W Thomas, Atri Rudra, and Christopher Re.
\newblock Hungry hungry hippos: Towards language modeling with state space models.
\newblock In \emph{The Eleventh International Conference on Learning Representations}, 2023.

\bibitem[Gao et~al.(2019)Gao, Ma, Zhao, Liu, and Yuille]{gao2019nddr}
Yuan Gao, Jiayi Ma, Mingbo Zhao, Wei Liu, and Alan~L Yuille.
\newblock Nddr-cnn: Layerwise feature fusing in multi-task cnns by neural discriminative dimensionality reduction.
\newblock In \emph{Proceedings of the IEEE/CVF conference on computer vision and pattern recognition}, pages 3205--3214, 2019.

\bibitem[Gu and Dao(2024)]{gu2024mamba}
Albert Gu and Tri Dao.
\newblock Mamba: Linear-time sequence modeling with selective state spaces, 2024.

\bibitem[Gu et~al.(2021{\natexlab{a}})Gu, Goel, and Ré]{DBLP:journals/corr/abs-2111-00396}
Albert Gu, Karan Goel, and Christopher Ré.
\newblock Efficiently modeling long sequences with structured state spaces.
\newblock \emph{CoRR}, abs/2111.00396, 2021{\natexlab{a}}.

\bibitem[Gu et~al.(2021{\natexlab{b}})Gu, Johnson, Goel, Saab, Dao, Rudra, and R{\'e}]{gu2021combining}
Albert Gu, Isys Johnson, Karan Goel, Khaled Saab, Tri Dao, Atri Rudra, and Christopher R{\'e}.
\newblock Combining recurrent, convolutional, and continuous-time models with linear state space layers.
\newblock \emph{Advances in neural information processing systems}, 34:\penalty0 572--585, 2021{\natexlab{b}}.

\bibitem[Huang et~al.(2024)Huang, Pei, You, Wang, Qian, and Xu]{huang2024localmamba}
Tao Huang, Xiaohuan Pei, Shan You, Fei Wang, Chen Qian, and Chang Xu.
\newblock Localmamba: Visual state space model with windowed selective scan.
\newblock \emph{arXiv preprint arXiv:2403.09338}, 2024.

\bibitem[Jeong and Yoon(2024)]{jeong2024quantifying}
Wooseong Jeong and Kuk-Jin Yoon.
\newblock Quantifying task priority for multi-task optimization.
\newblock In \emph{Proceedings of the IEEE/CVF Conference on Computer Vision and Pattern Recognition}, pages 363--372, 2024.

\bibitem[Kendall et~al.(2018)Kendall, Gal, and Cipolla]{kendall2018multi}
Alex Kendall, Yarin Gal, and Roberto Cipolla.
\newblock Multi-task learning using uncertainty to weigh losses for scene geometry and semantics.
\newblock In \emph{Proceedings of the IEEE conference on computer vision and pattern recognition}, pages 7482--7491, 2018.

\bibitem[Li et~al.(2024)Li, McDonagh, Leonardis, and Bilen]{li2024multitask}
Wei-Hong Li, Steven McDonagh, Ales Leonardis, and Hakan Bilen.
\newblock Multi-task learning with 3d-aware regularization.
\newblock In \emph{The Twelfth International Conference on Learning Representations}, 2024.

\bibitem[Lin et~al.(2024)Lin, Jiang, Chen, Zhang, Liu, and Chen]{lin2024mtmamba}
Baijiong Lin, Weisen Jiang, Pengguang Chen, Yu Zhang, Shu Liu, and Ying-Cong Chen.
\newblock {MTMamba}: Enhancing multi-task dense scene understanding by mamba-based decoders.
\newblock In \emph{European Conference on Computer Vision}, 2024.

\bibitem[Liu et~al.(2024)Liu, Tian, Zhao, Yu, Xie, Wang, Ye, and Liu]{DBLP:journals/corr/abs-2401-10166}
Yue Liu, Yunjie Tian, Yuzhong Zhao, Hongtian Yu, Lingxi Xie, Yaowei Wang, Qixiang Ye, and Yunfan Liu.
\newblock Vmamba: Visual state space model.
\newblock \emph{CoRR}, abs/2401.10166, 2024.

\bibitem[Liu et~al.(2021)Liu, Lin, Cao, Hu, Wei, Zhang, Lin, and Guo]{liu2021swin}
Ze Liu, Yutong Lin, Yue Cao, Han Hu, Yixuan Wei, Zheng Zhang, Stephen Lin, and Baining Guo.
\newblock Swin transformer: Hierarchical vision transformer using shifted windows.
\newblock In \emph{Proceedings of the IEEE/CVF international conference on computer vision}, pages 10012--10022, 2021.

\bibitem[Lu et~al.(2024)Lu, Cao, and Wang]{lu2024swiss}
Yuxiang Lu, Shengcao Cao, and Yu-Xiong Wang.
\newblock Swiss army knife: Synergizing biases in knowledge from vision foundation models for multi-task learning.
\newblock \emph{arXiv preprint arXiv:2410.14633}, 2024.

\bibitem[Maninis et~al.(2019)Maninis, Radosavovic, and Kokkinos]{maninis2019attentive}
Kevis-Kokitsi Maninis, Ilija Radosavovic, and Iasonas Kokkinos.
\newblock Attentive single-tasking of multiple tasks.
\newblock In \emph{Proceedings of the IEEE/CVF conference on computer vision and pattern recognition}, pages 1851--1860, 2019.

\bibitem[Mehta et~al.(2022)Mehta, Gupta, Cutkosky, and Neyshabur]{mehta2022long}
Harsh Mehta, Ankit Gupta, Ashok Cutkosky, and Behnam Neyshabur.
\newblock Long range language modeling via gated state spaces.
\newblock \emph{arXiv preprint arXiv:2206.13947}, 2022.

\bibitem[Misra et~al.(2016)Misra, Shrivastava, Gupta, and Hebert]{misra2016cross}
Ishan Misra, Abhinav Shrivastava, Abhinav Gupta, and Martial Hebert.
\newblock Cross-stitch networks for multi-task learning.
\newblock In \emph{Proceedings of the IEEE conference on computer vision and pattern recognition}, pages 3994--4003, 2016.

\bibitem[Navon et~al.(2022)Navon, Shamsian, Achituve, Maron, Kawaguchi, Chechik, and Fetaya]{navon2022multi}
Aviv Navon, Aviv Shamsian, Idan Achituve, Haggai Maron, Kenji Kawaguchi, Gal Chechik, and Ethan Fetaya.
\newblock Multi-task learning as a bargaining game.
\newblock \emph{arXiv preprint arXiv:2202.01017}, 2022.

\bibitem[Shoouri et~al.(2023)Shoouri, Yang, Fan, and Kim]{shoouri2023efficient}
Sara Shoouri, Mingyu Yang, Zichen Fan, and Hun-Seok Kim.
\newblock Efficient computation sharing for multi-task visual scene understanding.
\newblock In \emph{Proceedings of the IEEE/CVF International Conference on Computer Vision}, pages 17130--17141, 2023.

\bibitem[Silberman et~al.(2012)Silberman, Hoiem, Kohli, and Fergus]{silberman2012indoor}
Nathan Silberman, Derek Hoiem, Pushmeet Kohli, and Rob Fergus.
\newblock Indoor segmentation and support inference from rgbd images.
\newblock In \emph{Computer Vision--ECCV 2012: 12th European Conference on Computer Vision, Florence, Italy, October 7-13, 2012, Proceedings, Part V 12}, pages 746--760. Springer, 2012.

\bibitem[Sinodinos and Armanfard(2024)]{sinodinos2024ema}
Dimitrios Sinodinos and Narges Armanfard.
\newblock Ema-net: Efficient multitask affinity learning for dense scene predictions.
\newblock \emph{arXiv preprint arXiv:2401.11124}, 2024.

\bibitem[Smith et~al.(2022)Smith, Warrington, and Linderman]{DBLP:journals/corr/abs-2208-04933}
Jimmy T.~H. Smith, Andrew Warrington, and Scott~W. Linderman.
\newblock Simplified state space layers for sequence modeling.
\newblock \emph{CoRR}, abs/2208.04933, 2022.

\bibitem[Sun et~al.(2021)Sun, Probst, Paudel, Popovi{\'c}, Kanakis, Patel, Dai, and Van~Gool]{sun2021task}
Guolei Sun, Thomas Probst, Danda~Pani Paudel, Nikola Popovi{\'c}, Menelaos Kanakis, Jagruti Patel, Dengxin Dai, and Luc Van~Gool.
\newblock Task switching network for multi-task learning.
\newblock In \emph{Proceedings of the IEEE/CVF international conference on computer vision}, pages 8291--8300, 2021.

\bibitem[Vandenhende et~al.(2020)Vandenhende, Georgoulis, and Van~Gool]{vandenhende2020mti}
Simon Vandenhende, Stamatios Georgoulis, and Luc Van~Gool.
\newblock Mti-net: Multi-scale task interaction networks for multi-task learning.
\newblock In \emph{Computer Vision--ECCV 2020: 16th European Conference, Glasgow, UK, August 23--28, 2020, Proceedings, Part IV 16}, pages 527--543. Springer, 2020.

\bibitem[Vandenhende et~al.(2021)Vandenhende, Georgoulis, Van~Gansbeke, Proesmans, Dai, and Van~Gool]{vandenhende2021multi}
Simon Vandenhende, Stamatios Georgoulis, Wouter Van~Gansbeke, Marc Proesmans, Dengxin Dai, and Luc Van~Gool.
\newblock Multi-task learning for dense prediction tasks: A survey.
\newblock \emph{IEEE transactions on pattern analysis and machine intelligence}, 44\penalty0 (7):\penalty0 3614--3633, 2021.

\bibitem[Xu et~al.(2018)Xu, Ouyang, Wang, and Sebe]{xu2018pad}
Dan Xu, Wanli Ouyang, Xiaogang Wang, and Nicu Sebe.
\newblock Pad-net: Multi-tasks guided prediction-and-distillation network for simultaneous depth estimation and scene parsing.
\newblock In \emph{Proceedings of the IEEE conference on computer vision and pattern recognition}, pages 675--684, 2018.

\bibitem[Xu et~al.(2023)Xu, Li, Yuan, Yang, and Zhang]{xu2023multi}
Yangyang Xu, Xiangtai Li, Haobo Yuan, Yibo Yang, and Lefei Zhang.
\newblock Multi-task learning with multi-query transformer for dense prediction.
\newblock \emph{IEEE Transactions on Circuits and Systems for Video Technology}, 2023.

\bibitem[Yang et~al.(2024)Yang, Chen, Espinosa, Ericsson, Wang, Liu, and Crowley]{DBLP:journals/corr/abs-2403-17695}
Chenhongyi Yang, Zehui Chen, Miguel Espinosa, Linus Ericsson, Zhenyu Wang, Jiaming Liu, and Elliot~J. Crowley.
\newblock Plainmamba: Improving non-hierarchical mamba in visual recognition.
\newblock \emph{CoRR}, abs/2403.17695, 2024.

\bibitem[Ye et~al.(2024)Ye, Lyu, Wang, Zhang, and Tsang]{ye2024adaptive}
Feiyang Ye, Yueming Lyu, Xuehao Wang, Yu Zhang, and Ivor Tsang.
\newblock Adaptive stochastic gradient algorithm for black-box multi-objective learning.
\newblock In \emph{The Twelfth International Conference on Learning Representations}, 2024.

\bibitem[Ye and Xu(2022)]{ye2022inverted}
Hanrong Ye and Dan Xu.
\newblock Inverted pyramid multi-task transformer for dense scene understanding.
\newblock In \emph{European Conference on Computer Vision}, pages 514--530. Springer, 2022.

\bibitem[Ye and Xu(2023)]{ye2023taskprompter}
Hanrong Ye and Dan Xu.
\newblock Taskprompter: Spatial-channel multi-task prompting for dense scene understanding.
\newblock In \emph{The Eleventh International Conference on Learning Representations}, 2023.

\bibitem[Ye and Xu(2024)]{ye2024invpt++}
Hanrong Ye and Dan Xu.
\newblock Invpt++: Inverted pyramid multi-task transformer for visual scene understanding.
\newblock \emph{IEEE Transactions on Pattern Analysis and Machine Intelligence}, 2024.

\bibitem[Yu et~al.(2020)Yu, Kumar, Gupta, Levine, Hausman, and Finn]{yu2020gradient}
Tianhe Yu, Saurabh Kumar, Abhishek Gupta, Sergey Levine, Karol Hausman, and Chelsea Finn.
\newblock Gradient surgery for multi-task learning.
\newblock \emph{Advances in Neural Information Processing Systems}, 33:\penalty0 5824--5836, 2020.

\bibitem[Zhang et~al.(2023)Zhang, Fan, Ye, Zhang, Ye, Li, Cai, and Chen]{zhang2023rethinking}
Jingdong Zhang, Jiayuan Fan, Peng Ye, Bo Zhang, Hancheng Ye, Baopu Li, Yancheng Cai, and Tao Chen.
\newblock Rethinking of feature interaction for multi-task learning on dense prediction.
\newblock \emph{arXiv preprint arXiv:2312.13514}, 2023.

\bibitem[Zhang et~al.(2019)Zhang, Cui, Xu, Yan, Sebe, and Yang]{zhang2019pattern}
Zhenyu Zhang, Zhen Cui, Chunyan Xu, Yan Yan, Nicu Sebe, and Jian Yang.
\newblock Pattern-affinitive propagation across depth, surface normal and semantic segmentation.
\newblock In \emph{Proceedings of the IEEE/CVF conference on computer vision and pattern recognition}, pages 4106--4115, 2019.

\bibitem[Zhao et~al.(2024)Zhao, Chen, Zhang, Xiao, Bai, and Ouyang]{zhao2024rs}
Sijie Zhao, Hao Chen, Xueliang Zhang, Pengfeng Xiao, Lei Bai, and Wanli Ouyang.
\newblock Rs-mamba for large remote sensing image dense prediction.
\newblock \emph{arXiv preprint arXiv:2404.02668}, 2024.

\bibitem[Zhou et~al.(2020)Zhou, Cui, Xu, Zhang, Wang, Zhang, and Yang]{zhou2020pattern}
Ling Zhou, Zhen Cui, Chunyan Xu, Zhenyu Zhang, Chaoqun Wang, Tong Zhang, and Jian Yang.
\newblock Pattern-structure diffusion for multi-task learning.
\newblock In \emph{Proceedings of the IEEE/CVF conference on computer vision and pattern recognition}, pages 4514--4523, 2020.

\bibitem[Zhu et~al.(2024)Zhu, Liao, Zhang, Wang, Liu, and Wang]{DBLP:conf/icml/ZhuL0W0W24}
Lianghui Zhu, Bencheng Liao, Qian Zhang, Xinlong Wang, Wenyu Liu, and Xinggang Wang.
\newblock Vision mamba: Efficient visual representation learning with bidirectional state space model.
\newblock In \emph{ICML}, 2024.

\end{thebibliography}
}
\appendix

\clearpage
\setcounter{page}{1}
\maketitlesupplementary

\begin{appendices}

\section{Details of DBIM} 
\begin{figure}[ht]
    \centering
    \includegraphics[width=\linewidth]{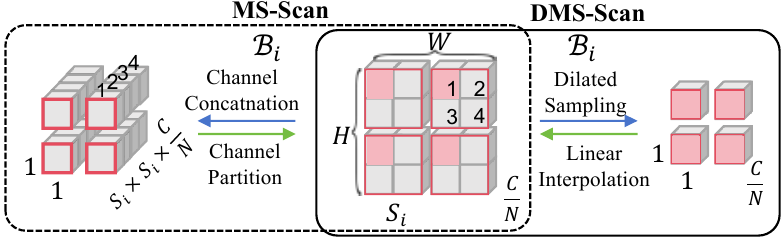}
    \caption{Comparison of MS-Scan and DMS-Scan.}
    \label{fig_dms}
\end{figure}

To further validate the effectiveness of our method, we present Dilated BIM (DBIM), a lightweight version of BIM, which achieves superior performance with reduced computational complexity and parameter count compared to MTMamba.
In DBIM, we replace MS-Scan with a more lightweight variant, DMS-Scan, which conducts sparse scanning within each scanning branch $\mathcal{B}$. Specifically, as shown in \cref{fig_dms}, we perform dilated sampling in generating multi-scale sequences from image features instead of using all tokens. When restoring sequences to image features, we perform linear interpolation. These operations do not introduce any parameters
and exhibit a reduced computational burden due to sampling a subset of tokens for modeling.

\section{More Ablation Studies}

\noindent \textbf{Effect of scan mode order in BI-Scan.}
We performed an ablation study to assess the impact of the scanning order in BI-Scan, containing two sequences: TF $\rightarrow$ PF (Task-First mode then Position-First mode) and PF $\rightarrow$ TF (the reverse). The results in \cref{orderofct} show that the TF $\rightarrow$ PF setting achieves better performance. Importantly, both combined strategies surpass the performance of using either TF or PF individually, which verifies the inherent complementarity between the two scanning modes.

\begin{table}[ht]
\setlength{\tabcolsep}{2pt}
\small
\centering
\caption{Effect of scan mode order in BI-Scan.}
\label{orderofct}
\resizebox{\linewidth}{!}{
\begin{tabular}{c |c c c c| c c}
\toprule
\multirow{2}{*}{\textbf{Setting}}& \textbf{Semseg}& \textbf{Depth}& \textbf{Normal}& \textbf{Boundary} &\textbf{FLOPs}& \textbf{\# Params}\\ 

& mIoU $\uparrow$ & RMSE$\downarrow$ & mErr $\downarrow$ & odsF $\uparrow$ & (G) $\downarrow$ & (M) $\downarrow$\\
\midrule
TF & 56.36	&0.4905	&18.67	&78.70	&510&289	 \\
PF &56.96&	0.4883	&18.68	&78.70&510&	289 \\
TF $\rightarrow$ PF&\textbf{57.11}&0.4856&\textbf{18.66}&\textbf{78.80}&547&290\\
PF $\rightarrow$ TF&56.55	&\textbf{0.4806}	&18.71	&78.70 &547&290\\
 
\bottomrule
\end{tabular}
}
\end{table}



\noindent \textbf{Model efficiency with varying task quantities.}
To further validate the linear complexity of our method, we conducted an experiment on the NYUD dataset with a varying number of tasks. Specifically, we benchmarked our model's performance using 2, 3, and 4 tasks. The results, presented in \cref{num}, show that as the number of tasks increases, the incremental computational cost (GFLOPs) and the number of additional parameters both remain constant for each new task. This observation empirically verifies that our model's complexity scales linearly with the number of tasks.

\begin{table}[ht]
\setlength{\tabcolsep}{2pt}
\small
\centering
\caption{Model efficiency with varying task quantities.}
\label{num}
\resizebox{\linewidth}{!}{
\begin{tabular}{c |c c c c| c c}
\toprule
\multirow{2}{*}{\textbf{Methods}}& \textbf{Semseg}& \textbf{Depth}& \textbf{Normal}& \textbf{Boundary} &\textbf{FLOPs}& \textbf{\# Params}\\ 
& mIoU $\uparrow$ & RMSE$\downarrow$ & mErr $\downarrow$ & odsF $\uparrow$ & (G) $\downarrow$ & (M) $\downarrow$\\
\midrule
2 tasks& 57.20&0.4715&-& -&375&304\\
3 tasks&55.91&0.4891&18.63&-&461(\textbf{+86})&346(\textbf{+42})\\
4 tasks&57.40& 0.4733&18.55&78.72 & 547(\textbf{+86})&388(\textbf{+42})\\

\bottomrule
\end{tabular}
}
\end{table}

\section{More Visual Comparison Results}
\begin{figure}[h]
    \centering
    \includegraphics[width=\linewidth]{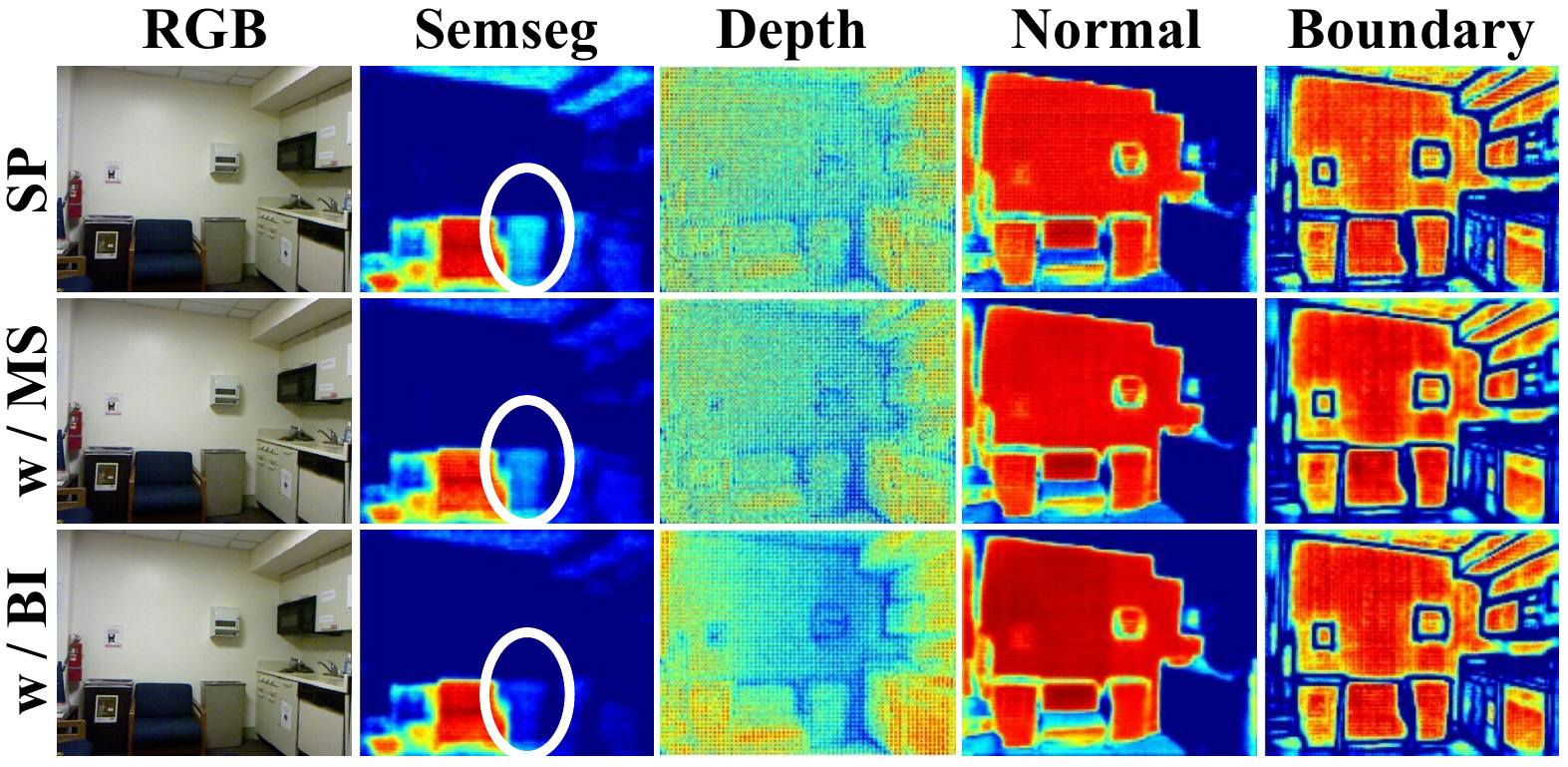}
    \caption{Effect of BI-Scan and MS-Scan on attention patterns.}
    \label{ms-bi}
\end{figure}
\noindent \textbf{Effect of BI-Scan and MS-Scan on Task Attention Patterns.}
To systematically validate the efficacy of our proposed scanning mechanisms in task representation enhancement, we conduct a quantitative visual analysis of attention patterns in the final MFR block, as illustrated in Figure \ref{ms-bi}. Our comparative investigation examines three distinct feature configurations: (a) Task-specific features (\textbf{SP}), (b) Features enhanced by MS-Scan in MSST block (\textbf{W/ MS}), and (c) Features refined via the BI-Scan in BCFR block (\textbf{W/ BI}). The results reveal two critical insights: First, the MS-Scan mechanism substantially enriches task-specific feature details through multi-scale contextual integration.
Second, the BI-Scan induces task-aligned attention redistribution, effectively suppressing irrelevant spatial responses while amplifying task-critical regions. This synergistic effect is demonstrated by the progressive attention focusing observed in white-circled areas of \cref{ms-bi}.

\begin{figure}[ht]
    \centering
    \includegraphics[width=0.95\linewidth]{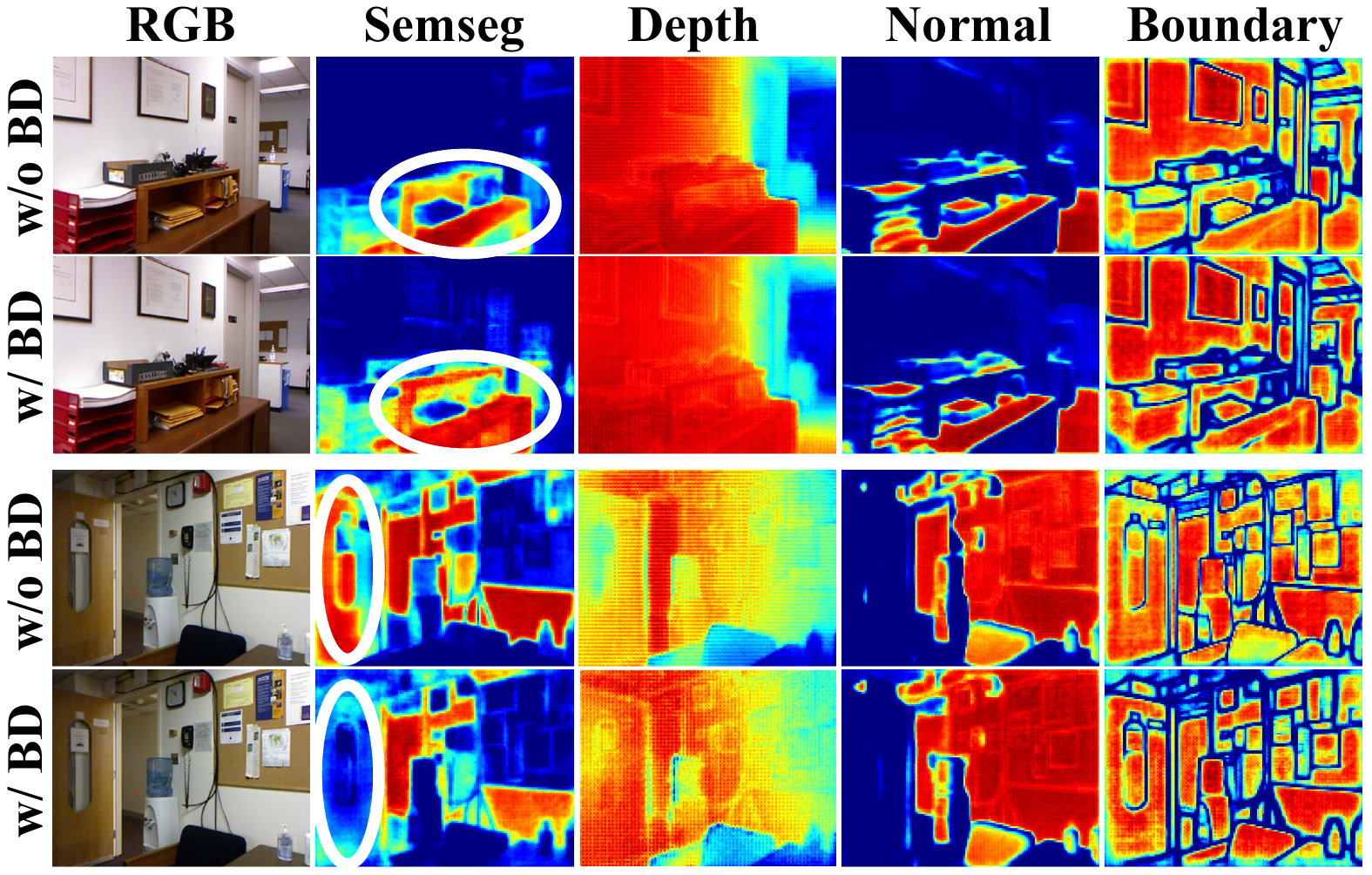}
    \caption{Effect of bidirectional scan on attention patterns.}
    \label{bi}
\end{figure}

\noindent \textbf{Effect of Bidirectional Scan on Task Attention Patterns.}
To validate the bidirectional scan efficacy in the BI-Scan mechanism, we conduct a controlled comparative study of task attention patterns in the final MFR block under two experimental configurations: (1) Unidirectional modeling (\textbf{w/o BD}) with forward scanning only, and (2) Bidirectional modeling (\textbf{w/ BD}), while maintaining identical feature channel dimensions for fair comparison. As shown by the white-circled regions in \cref{bi}, in the upper panel exemplars, bidirectional scanning exhibits enhanced attention localization in task-critical regions while preserving the structural integrity of target areas. The lower panel reveals the model without bidirectional scanning, which erroneously groups architecturally distinct elements (doors/walls) due to a flawed understanding of the scene structure. whereas bidirectional implementation achieves precise separation through detailed and comprehensive cross-task interactions.



\begin{figure}[ht]
    \centering
    \includegraphics[width=\linewidth]{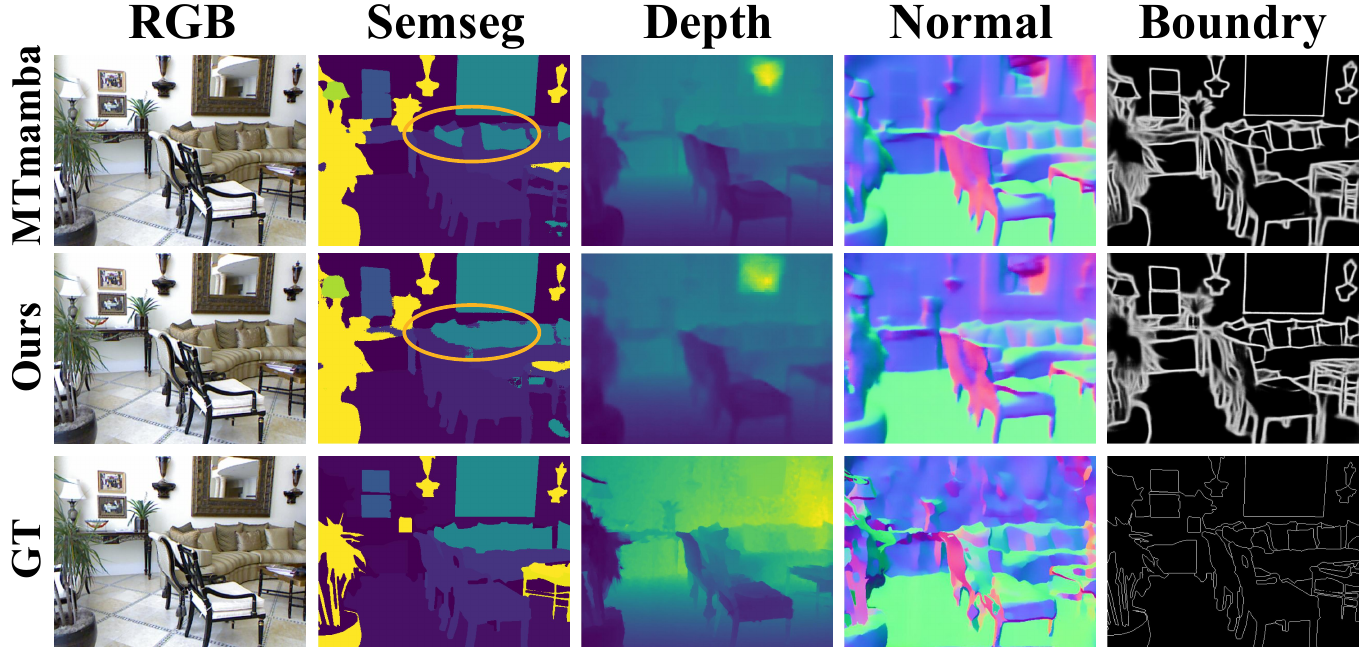}
    \caption{More qualitative comparison on NYUD-v2.}
    \label{psv1}
\end{figure}

\begin{figure}[t]
    \centering
    \includegraphics[width=\linewidth]{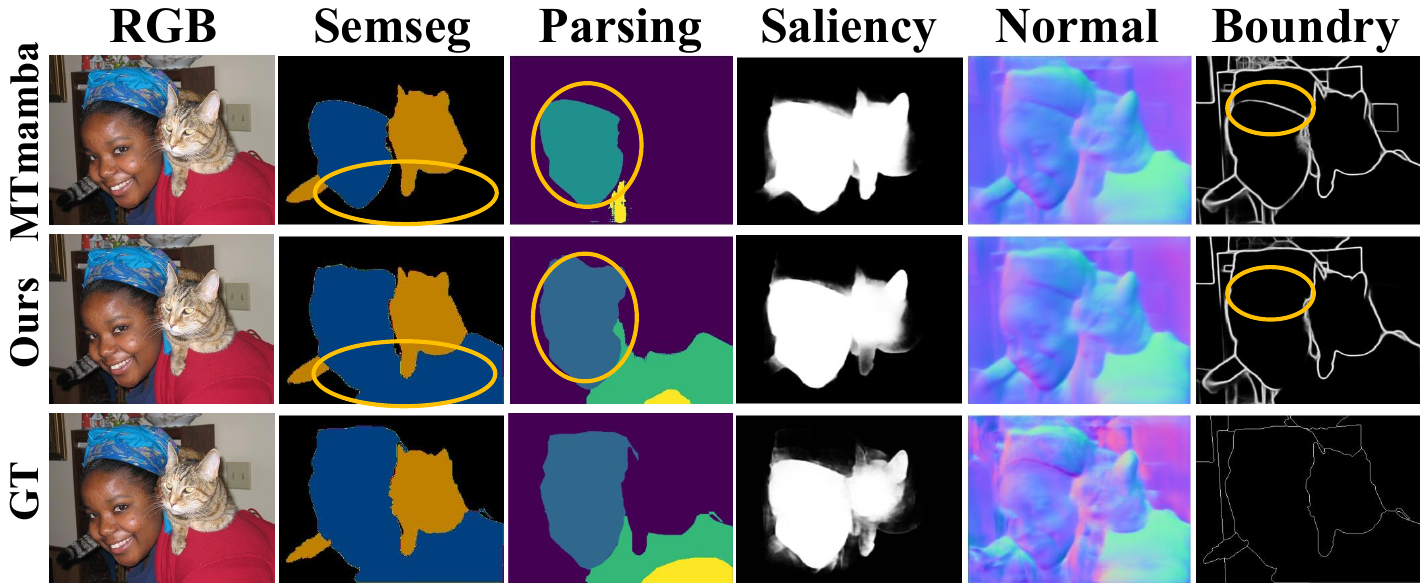}
 \caption{More qualitative comparison on Pascal-Context.}
     \label{psv2}
    
\end{figure}

\begin{figure}[t]
    \centering
    \includegraphics[width=\linewidth]{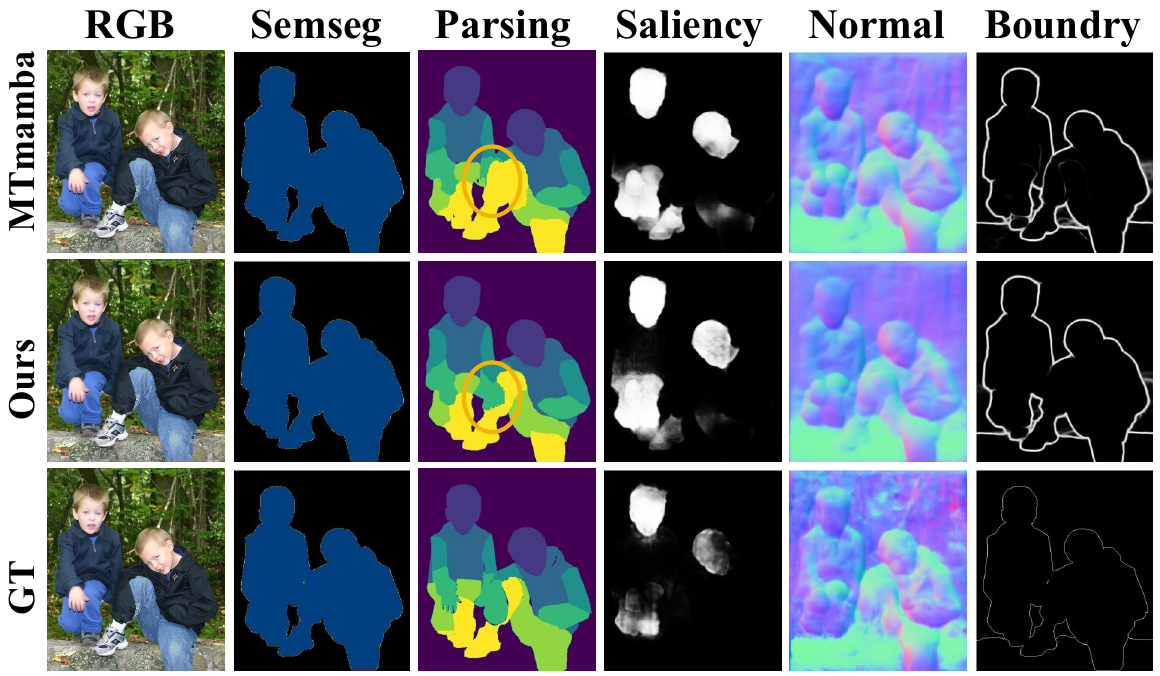}
    \caption{More qualitative comparison on Pascal-Context.}
    \label{psv3}
\end{figure}

\noindent \textbf{Qualitative Comparison with state-of-the-art method.}
We present more qualitative results compared with the SOTA methods. In \cref{psv1,psv2,psv3}, the results indicate that our method generates more detailed multi-task predictions, as highlighted in the circled regions.

\null 


   
        
    

       



\end{appendices}


\end{document}